\documentclass[sigconf]{acmart}

\usepackage{booktabs} 
\usepackage{multirow}
\usepackage{placeins}
\usepackage{caption}
\usepackage{pifont} 

\providecommand{\OASISBibFile}{main}

\newcommand{\eg}{\textit{e.g.}}
\newcommand{\ie}{\textit{i.e.}}

\copyrightyear{2026}
\acmYear{2026}
\setcopyright{cc}
\setcctype{by}
\acmConference[MM '26]{Proceedings of the 34th ACM International Conference on Multimedia}{November 10--14, 2026}{Rio de Janeiro, Brazil}
\acmBooktitle{Proceedings of the 34th ACM International Conference on Multimedia (MM '26), November 10--14, 2026, Rio de Janeiro, Brazil}
\acmDOI{10.1145/3767308.3835228}
\acmISBN{979-8-4007-2213-4/2026/11}
\begin{document}

\title{OASIS: Online Anchor-centric Scene Inference System for Open-Vocabulary Retrieval and Navigation in Dynamic Environments}

\renewcommand{\shorttitle}{OASIS: Online Anchor-centric Scene Inference System}


\author{Kaiyao Liao}
\affiliation{%
  \institution{Beihang University}
  \city{Beijing}
  \country{China}
}
\email{2406104@buaa.edu.cn}

\author{Qingfeng Li}
\affiliation{%
  \institution{Beihang University}
  \city{Beijing}
  \country{China}
}
\email{liqingfeng@buaa.edu.cn}

\author{Chen Chen}
\affiliation{%
  \institution{Hangzhou Innovation Institute of Beihang University}
  \city{Hangzhou}
  \country{China}
}
\email{chenx2ovo@buaa.edu.cn}

\author{Xiaoze Wu}
\affiliation{%
  \institution{Beihang University}
  \city{Beijing}
  \country{China}
}
\email{wwwww@buaa.edu.cn}

\author{Qing Sun}
\affiliation{%
  \institution{Hangzhou Innovation Institute of Beihang University}
  \city{Hangzhou}
  \country{China}
}
\email{hollysun@buaa.edu.cn}

\author{Jianwei Niu}
\authornote{Corresponding author.}
\affiliation{%
  \department{School of Computer Science and Engineering}
  \institution{Beihang University}
  \city{Beijing}
  \country{China}
}
\email{niujianwei@buaa.edu.cn}

\renewcommand{\shortauthors}{Kaiyao Liao et al.}

\begin{abstract}
Reliable robotic mapping in dynamic real-world environments is essential for embodied intelligence and long-term autonomy. However, existing methods often involve a trade-off: lightweight online mapping is efficient but captures limited semantic context, while dense mapping provides richer semantics at higher computational cost and is more vulnerable to distractors. They also struggle to relocalize objects after scene changes. To address these challenges, we propose the Online Anchor-centric Scene Inference System (OASIS), a unified system for online mapping, complex open-vocabulary retrieval, and navigation in dynamic environments. Motivated by the observation that indoor scenes are organized around stable semantic anchors, OASIS represents the environment using a sparse set of anchor nodes, each maintaining a unified multimodal representation that integrates semantic descriptions, visual features, and geometric states for reasoning, verification, and navigation. Its compact anchor-centric memory supports on-demand evidence extraction and hierarchical retrieval, reducing redundant storage while improving retrieval reliability. To handle scene dynamics, OASIS further reuses transient observations collected during navigation to relocalize moved objects. Experiments in both simulation and real-world environments, including standard retrieval benchmarks and navigation tasks under both static and dynamic settings, demonstrate that our system can perform complex open-vocabulary retrieval and navigation, while also enabling relocation of displaced objects after scene changes.
\end{abstract}


\maketitle

\begin{figure*}[t]
    \centering
    \includegraphics[width=\textwidth]{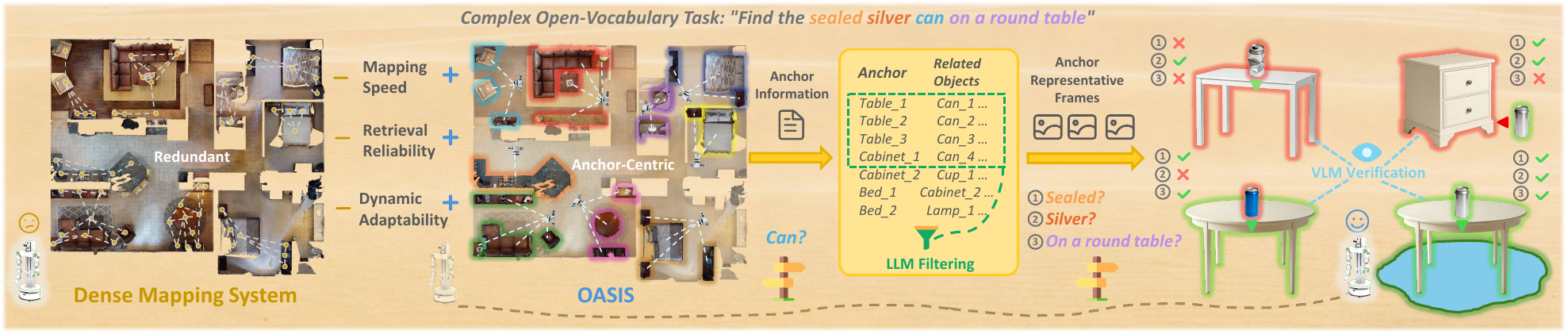}
    \caption{\textbf{Comparison of retrieval paradigms for complex open-vocabulary queries.} Unlike traditional dense mapping (left) which struggles with redundant context and visual distractors, our OASIS (right) constructs a lightweight anchor-centric memory. By leveraging an LLM to filter irrelevant regions and performing on-demand VLM verification, OASIS efficiently isolates the target based on fine-grained attributes and spatial relationships, significantly improving retrieval reliability.}
    \label{fig:oasis_teaser}
\end{figure*}


\section{Introduction}
\label{sec:intro}
Modern embodied agents face increasingly complex tasks in unstructured environments, where instructions involve rich, open-vocabulary descriptions beyond simple object categories. A user might command a robot to ``find the sealed silver can on a round table'' or ``locate the bed with a lit lamp.'' Fulfilling such requests requires a scene representation that transcends traditional geometric mapping, demanding fine-grained attribute recognition (\eg, color, material, state) and spatial relationship reasoning. Crucially, these capabilities must remain robust in dynamic real-world settings where objects are frequently repositioned. Consequently, the challenge has shifted from static mapping to constructing a dynamic, semantic-rich scene graph capable of supporting intricate queries and adapting to environmental changes on the fly.

However, deploying semantic maps in real-world systems still involves a practical trade-off between efficiency and reasoning capability. Dense mapping approaches~\cite{dense-1,dense-2,dense-3} embed rich semantics into many spatial elements, providing detailed understanding but sometimes introducing clutter and distractors, while also requiring relatively high computational and memory resources. Conversely, lightweight methods~\cite{light-1,light-2} are more efficient, but may not preserve enough spatial context for complex reasoning, making it harder to distinguish visually similar objects based on their surroundings or attributes. A remaining challenge appears in dynamic environments. Although some methods address object relocalization, they often depend on repeated re-observation. During active navigation, rapid robot motion and a limited camera view can cause shifted targets to be missed, which reduces the efficiency of these scanning-based strategies.

To bridge this gap, we propose the Online Anchor-centric Scene Inference System (OASIS) for robust open-vocabulary retrieval and dynamic object relocalization. As illustrated in Figure~\ref{fig:oasis_teaser}, OASIS avoids processing redundant context globally through an on-demand retrieval strategy. Our key insight is that indoor environments are inherently structured around stable, large-scale objects (\eg, beds, tables, sofas), which we define as semantic anchors. Inspired by human episodic memory, OASIS constructs an anchor-centric graph rather than a dense per-pixel map, retaining only salient semantic descriptions and high-quality representative frames. During retrieval, a Large Language Model (LLM) first reasons over the semantic anchors to filter irrelevant regions, followed by on-demand Vision-Language Model (VLM) verification of the retrieved representative frames. This strategy keeps memory lightweight while preserving the ability to distinguish fine-grained visual attributes. In dynamic settings, OASIS additionally maintains transient observations along the navigation trajectory in a lightweight buffer; if a target is missing at its initial anchor, the system queries this buffer and re-verifies candidate views for reliable relocalization.

Our main contributions are summarized as follows:
\begin{itemize}
    \item We present OASIS, an \textbf{online anchor-centric scene inference system} that unifies online mapping, open-vocabulary retrieval, and navigation in dynamic environments.

    \item We propose a human-inspired \textbf{episodic memory curation mechanism} that uses scale-aware scoring and spatio-semantic clustering to select representative frames, yielding a multimodal \textbf{anchor-centric scene graph} with integrated semantic, visual, and geometric representations.

    \item Built upon this scene graph, OASIS supports \textbf{hierarchical coarse-to-fine retrieval} and view-driven navigation, and introduces a \textbf{Transient Visual Buffer} to reliably relocalize moved objects after scene changes.

    \item We conduct extensive simulation and real-world experiments on retrieval and navigation under static and dynamic settings, demonstrating strong performance and reliable object relocation after scene changes.
\end{itemize}

\section{Related Work}

\subsection{Open-Vocabulary 3D Scene Understanding}
Closed-set semantic mapping relies on detectors trained for fixed object categories~\cite{close-1,close-2,close-3,close-4}. Open-vocabulary methods instead distill features from vision-language models (\eg, CLIP~\cite{CLIP}) into 3D representations. Dense mapping methods~\cite{dense-1,dense-2,dense-3} project pixel-aligned features into 3D space, enabling fine-grained queries but incurring high computational and memory costs. Conversely, object-centric or lightweight semantic point cloud methods~\cite{light-1,light-2} construct sparse maps by aggregating features at the instance level. While these approaches are significantly more efficient, they often discard the surrounding visual context required for complex reasoning. Our work builds upon the efficiency of lightweight representations but augments them with an implicit storage mechanism to retain rich contextual details without the overhead of dense fields.

\subsection{Retrieval and Reasoning on Scene Graphs}
Recent advancements in open-vocabulary scene understanding have focused on reasoning about and locating objects within complex environments using foundation models. 
Early approaches primarily utilized CLIP-based matching~\cite{CLIP-1,CLIP-2,CLIP-3,dense-1}, which directly computes cosine similarity between text queries and stored visual features. While efficient for simple identification, these methods lack the capability for multi-step reasoning or handling spatial relationships.
To overcome these limitations, LLM-based methods~\cite{LLM-1,LLM-2,LLM-3,LLM-4,dense-2} leverage the reasoning power of LLMs to infer potential locations or generate navigation plans based on textual abstractions of the scene. However, relying solely on text descriptions often results in the loss of critical visual details (\eg, object states or specific textures) necessary for precise identification.
More recently, VLM-based approaches~\cite{dense-3,VLM-1} have emerged, directly processing images to retain visual fidelity. Despite their superior understanding, these methods typically suffer from high latency and computational costs, making them less practical for real-time applications.
Our work bridges these gaps by introducing an efficient VLM-driven mechanism that operates on implicit storage, enabling detailed visual reasoning without the heavy overhead of full-image processing.

\subsection{Object Navigation}
Object navigation requires agents to locate and move towards specific target instances within an environment~\cite{nav-1,nav-3,nav-4,nav-5,nav-6,nav-7,nav-8,nav-9,nav-10,nav-11}. Traditional approaches typically rely on static semantic maps, assuming that objects remain fixed throughout the mission. These methods combine SLAM with semantic segmentation to build a persistent representation, allowing the agent to plan paths to previously observed locations. 
However, real-world environments are inherently dynamic. Existing works on dynamic navigation~\cite{CLIP-1,light-2} typically address this by relying on continuous fine-grained modeling and geometric matching of objects along the trajectory to update the map. This strategy faces significant challenges: navigation often involves transient observations, making it difficult to capture and update the state of small objects accurately in real-time. Furthermore, when a target is missing from its expected location, these methods often lack the active reasoning capability to deduce its whereabouts. 
In contrast, our proposed method leverages subtle visual details observed along the path, enabling the system to infer the new location of displaced objects based on semantic context, even if the new location lies outside the immediate navigation trajectory.

\section{Method}
\begin{figure*}[t]
  \centering
  \includegraphics[width=\textwidth]{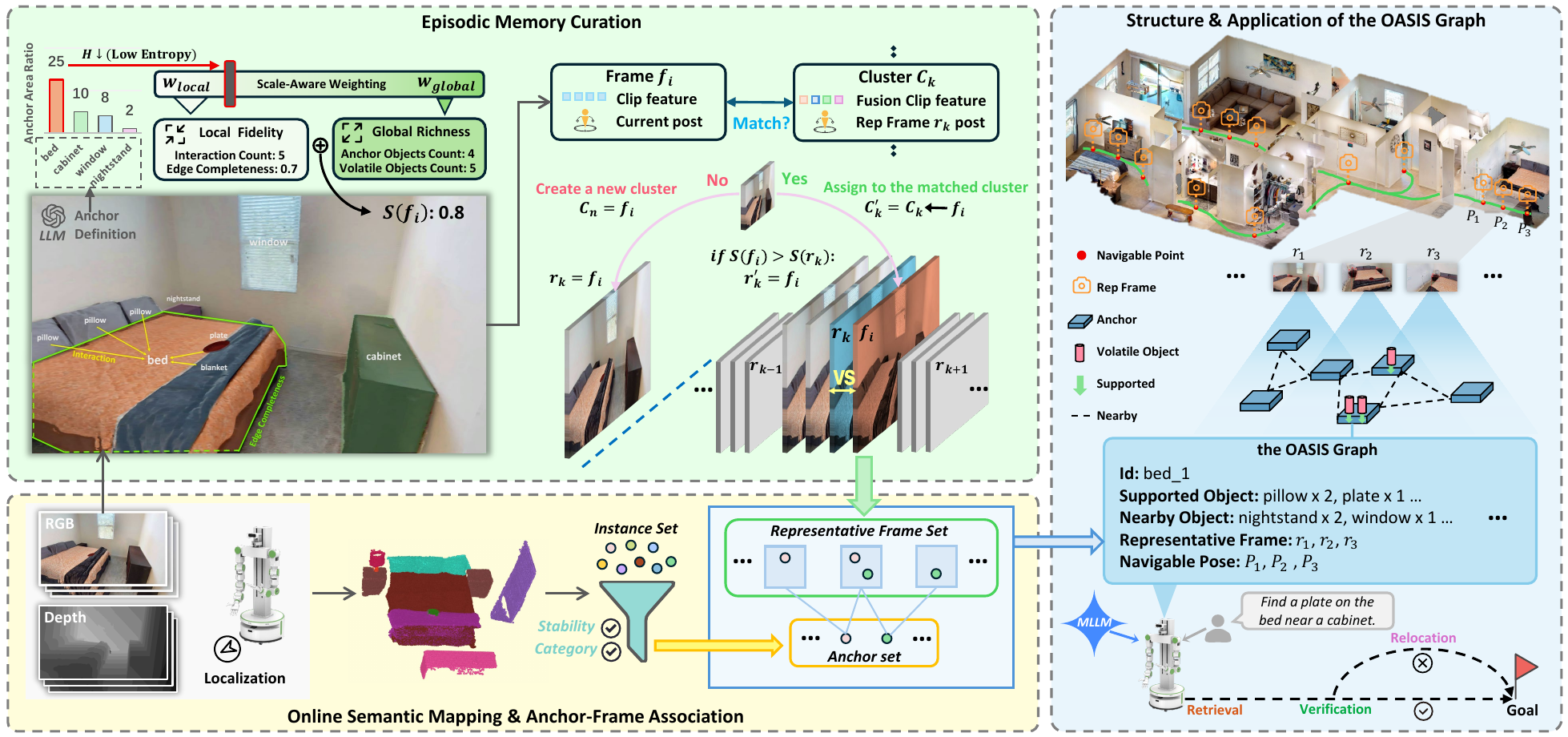}
  \caption{\textbf{System architecture of OASIS.} Given an RGB-D stream, the system concurrently performs episodic memory curation (top left) and online semantic mapping (bottom left) to extract 3D instances. To build the episodic memory, it utilizes scale-aware scoring and spatial-visual cluster assignment to continuously update and retain high-quality representative frames. By assigning these selected frames to their corresponding stable anchor instances, the system constructs the unified multimodal \textbf{OASIS Graph}, structured as shown on the right. Leveraging this graph alongside multimodal foundation models (LLM/VLM), the agent effectively executes hierarchical retrieval and robust view-driven navigation in dynamic environments.}
\label{fig:pipeline}
\end{figure*}

\subsection{Overview}
As illustrated in Figure~\ref{fig:pipeline}, OASIS addresses the challenge of simultaneous online mapping and open-vocabulary retrieval in dynamic environments. From the RGB-D stream and robot localization, it first performs online semantic mapping~\cite{light-2} to maintain an instance set, identify stable anchors, and associate observations with anchors; in parallel, it curates a compact episodic memory through spatial-visual clustering and scale-aware scoring that adaptively balances local fidelity and global richness to keep only high-value representative frames; finally, it fuses anchors, representative frames, structural relations, and navigable poses into the OASIS graph for coarse-to-fine candidate retrieval, VLM-based verification, and navigation, while a relocation branch handles target relocalization when objects are moved. For full reproducibility, exhaustive implementation details—including complete LLM/VLM prompts, exact mathematical formulations, and hyperparameter settings—are provided in Appendices~\ref{supp:prompt}--\ref{supp:prompt_screenshots}.


\subsection{Episodic Memory Curation}
\label{sec:memory_update}

Inspired by the human cognitive process of episodic memory, which retains only significant events and high-quality visual snapshots rather than continuous redundant streams, we design a dynamic memory curation mechanism. This module processes the input frame sequence $\mathcal{F} = \{f_1, f_2, \dots\}$ to maintain a set of spatial-visual clusters $\mathcal{C} = \{C_1, \dots, C_N\}$, where each cluster $C_k$ is represented by a single optimal frame $r_k$.

\textbf{LLM-Driven Anchor Definition.} 
Prior to memory curation, we explicitly define the set of target objects—termed \textit{Anchor Objects}—to align with human spatial cognition. Instead of manual selection, we leverage an LLM to identify object categories that satisfy three key criteria: (1) Volumetric Prominence (occupying significant space), (2) Stationarity (structurally static or heavy), and (3) Associative Utility (serving as natural landmarks for referencing nearby items). This LLM-driven approach ensures our system can generalize across diverse indoor environments by focusing on semantically stable landmarks (\eg, \textit{bed}, \textit{sofa}) while filtering out less informative observations to keep the representation focused.

\textbf{Scale-Aware Frame Scoring.} 
To retain only informative observations, we score each incoming frame by combining global scene richness and local anchor quality:
\begin{equation}
    S(f_i) = w_{\text{global}} S_{\text{global}}(f_i) + w_{\text{local}} S_{\text{local}}(f_i),
    \label{eq:score_total}
\end{equation}
\begin{equation}
    S_{\text{global}}(f_i) \propto
    \alpha \min(M, A_{\max}) +
    (1-\alpha)\min(N_{\text{obj}}, O_{\max}),
    \label{eq:score_global}
\end{equation}
\begin{equation}
    S_{\text{local}}(f_i) \propto
    \gamma \max_j(V_j) +
    (1-\gamma)\frac{1}{M}\sum_j V_j,
    \label{eq:score_local}
\end{equation}
\begin{equation}
    V_j =
    \beta E_j +
    (1-\beta)\min(C_j, C_{\max}),
    \label{eq:score_anchor}
\end{equation}
Here, $M$ and $N_{\text{obj}}$ denote the numbers of visible anchors and non-anchor objects in frame $f_i$, respectively. $A_{\max}$, $O_{\max}$, and $C_{\max}$ are clipping bounds for anchor count, object count, and interaction count. For each anchor $j$, $E_j \in [0,1]$ denotes edge completeness and $C_j$ denotes interaction count. The coefficients $\alpha,\beta,\gamma \in [0,1]$ balance the corresponding terms. Both $S_{\text{global}}$ and $S_{\text{local}}$ are normalized to $[0,1]$; when no anchor is visible ($M=0$), we set $S_{\text{local}}=0$.

The adaptive weights are determined by anchor coverage and anchor-area balance. Let $a_j$ denote the 2D image area of anchor $j$ in frame $f_i$, and define the normalized area distribution as $p_j = a_j / \sum_m a_m$, with entropy $H(f_i) = -\sum_j p_j \log p_j$, and the anchor coverage ratio as $r(f_i) = \sum_j a_j / A_{\text{frame}} \in [0,1]$, where $A_{\text{frame}}$ is the image area. We then compute
\begin{equation}
    w_{\text{global}} =
    \begin{cases}
    1-r(f_i), & M\le 1,\\
    (1-\lambda)(1-r(f_i)) + \lambda\,\dfrac{H(f_i)}{\log M}, & M>1,
    \end{cases}
\end{equation}
and set $w_{\text{local}} = 1 - w_{\text{global}}$, where $\lambda \in [0,1]$ balances coverage and entropy. Intuitively, frames dominated by a few large anchors emphasize local fidelity, while frames with smaller or more balanced anchors emphasize global context.

\textbf{Cluster Assignment and Representative Frame Update.} 
Armed with the frame score $S(f_i)$, we aim to curate a compact yet expressive memory from the highly redundant continuous RGB-D stream. We maintain the memory as a set of spatial-visual clusters. Each cluster $C_k$ is characterized by an aggregated visual feature $v_{C_k}$ and a \textit{representative frame} $r_k$—the highest-scoring frame that best encapsulates the visual and semantic information of that cluster's viewpoint.

To process the incoming stream, we first assign $f_i$ to an existing cluster. We extract its visual feature vector $v_i$ using a pre-trained CLIP encoder and obtain the camera pose $p_i$ from the localization module. A frame $f_i$ is assigned to $C_k$ if it satisfies both visual and spatial proximity:
\begin{equation}
    \text{Sim}_{\text{vis}}(v_i, v_{C_k}) > \tau_{\text{vis}} \quad \text{and} \quad \text{Sim}_{\text{spa}}(p_i, p_{r_k}) > \tau_{\text{spa}},
    \label{eq:clustering}
\end{equation}
where $\text{Sim}_{\text{vis}}$ denotes the cosine similarity, and $p_{r_k}$ is the pose of the cluster's representative frame $r_k$. 

If $f_i$ matches $C_k$, the cluster is updated in two ways. First, the cluster feature $v_{C_k}$ is dynamically updated via a weighted fusion parameterized by the frame scores $S(\cdot)$, ensuring that higher-quality frames contribute more significantly to the cluster representation. Second, we update the representative frame $r_k$ strictly when the new frame offers superior information density:
\begin{equation}
    r_k \leftarrow f_i \quad \text{if} \quad S(f_i) > S(r_k).
    \label{eq:update_policy}
\end{equation}
Conversely, if $f_i$ presents novel visual content or a distinct viewpoint (failing Equation \ref{eq:clustering}), it initializes a new cluster $C_{n}$ with itself as the initial representative frame ($r_n = f_i$), thereby progressively expanding the system's spatial-visual coverage. In our implementation, we set $\tau_{\text{vis}} = 0.8$, and define the spatial proximity $\tau_{\text{spa}}$ using translation and rotation bounds of $2\,\mathrm{m}$ and $45^{\circ}$, respectively.

\subsection{Anchor-Centric Scene Graph Construction}
\label{sec:mac_sg}

Building upon episodic memory, we construct the Anchor-Centric Scene Graph (the OASIS graph). This process integrates dense 3D information from the geometric mapping stream with high-quality visual snapshots from the memory stream into a unified multimodal representation.

\textbf{Online Semantic Mapping and Anchor-Frame Association.} 
Concurrently with memory curation, we leverage an online semantic mapping process~\cite{light-2}. Let $o_i \in \mathcal{O}$ denote an observed instance. The mapping process retains stable instances $o_i^{\mathrm{st}} \in \mathcal{O}^{\mathrm{st}}$ that persist for more than $n$ consecutive frames, and provides instance-level 3D boxes, point-cloud support, CLIP features, and high-confidence vertical relations (\textit{supported}) from box overlap. We then use the anchor categories defined in Sec.~\ref{sec:memory_update} to obtain anchor instances $A_j$ from $\mathcal{O}^{\mathrm{st}}$, and assign representative frames by intersecting each anchor's visible-frame set with memory frames $\mathcal{R}$. For compactness, if an anchor $A_j$ is linked to more than $K$ candidates, we keep the top-$K$ frames (default $K=5$) ranked by $S(\cdot)$.

\textbf{Graph Formalization.}
Formally, we define the OASIS graph as a graph $\mathcal{G} = \{\mathcal{V}, \mathcal{E}\}$, where $\mathcal{V}$ denotes the set of stable anchor instances. Unlike traditional scene graphs that attempt to exhaustively classify all relationships into static text labels, our OASIS graph functions as a multimodal index, prioritizing high-confidence geometric links while preserving rich visual contexts. The edges associated with an anchor $A_i$ are defined as:
\begin{itemize}
    \item \textbf{Geometric Edges:} We define directed edges $O_j \xrightarrow{\text{on}} A_i$ for objects physically supported by anchor $A_i$, derived from 3D bounding-box analysis.
    \item \textbf{Visual Edges:} Each anchor connects to its representative frames $\mathcal{R}_i = \{r_{i,1}, \dots, r_{i,K}\}$; stable instances that co-occur in these frames may exhibit diverse relative layouts (e.g., left, right, front, or behind), which we abstractly define as \textit{nearby} and use to form the anchor's \textit{nearby-list} for subsequent retrieval and verification.
\end{itemize}
In our implementation, the graph is serialized in JSON format. To maintain a lightweight structure, the visual edges store the file paths of the representative frames rather than raw image data. Additionally, each frame entry records its corresponding 6-DoF camera pose, which is utilized in subsequent navigation steps.

\subsection{Hierarchical Object Retrieval}
\label{sec:retrieval}

Leveraging the structured OASIS graph, we propose a three-stage \textit{coarse-to-fine} retrieval algorithm. This approach avoids the computational cost of scanning the entire scene by prioritizing semantically relevant anchors before performing detailed visual verification.

\textbf{Stage 1: Semantic Anchor Proposal (LLM-based).} 
Given a user query $\mathcal{Q}$ (\eg, ``\textit{Find the mug near the laptop}''), the system first reasons about the most probable locations using only the textual layer of the OASIS graph. An LLM functions as a planner, taking the query and the global anchor list as input. For each anchor $A_i$, the LLM analyzes its semantic category and the summary of its related objects (both \textit{supported-list} and \textit{nearby-list}). The LLM outputs a ranked list of candidate anchors $\mathcal{A}_{cand}$ (Top-3 by default) where the target is most likely to appear. This step purely relies on textual reasoning, effectively pruning the search space without processing any images.

\textbf{Stage 2: Visual Verification with Cross-Anchor Reasoning (VLM-based).} 
The system iterates through candidate anchors in $\mathcal{A}_{cand}$. For each queried anchor $A_i$, it retrieves representative frames and projects 2D boxes with IDs for \textit{all visible anchors} (not only $A_i$), enabling cross-anchor disambiguation in shared views where multiple anchors co-occur. The VLM then verifies target presence for query $\mathcal{Q}$, identifies the associated anchor $A_{selected}$ (which may differ from $A_i$), and outputs the selected anchor ID, the target relation to $A_{selected}$ (\ie, \texttt{self}, \texttt{on}, or \texttt{nearby}), and a detailed visual description with predicted class for subsequent instance matching.

\textbf{Stage 3: Precise Instance Localization (CLIP-based).} 
Once the VLM confirms a detection and identifies the reference anchor $A_{selected}$, we map the high-level perception back to a specific 3D instance UID ($u_{target}$). When the returned relation is \texttt{self}, the UID of $A_{selected}$ is directly returned. Otherwise, for relations \texttt{on} or \texttt{nearby}, we retrieve the corresponding candidate list (the \textit{supported-list} or \textit{nearby-list}) from node $A_{selected}$ and first filter candidates by the VLM-predicted class. When class-matched candidates exist, ambiguity (\eg, multiple mugs) is resolved by cosine similarity between candidate CLIP embeddings and the text embedding of the VLM-generated \textit{Visual Description}; when no class match exists (\eg, open-vocabulary mismatch), we compute CLIP similarity over the full candidate list and select the highest-scoring instance as the final target.

\textbf{Global Fallback Mechanism.} 
In cases where the hierarchical search fails (\ie, the target is not found in any candidate anchor's view), the system triggers a global fallback. We perform a global CLIP-based search across all scene instances using the visual description generated during the failure analysis (or the original query). To ensure efficiency, instances already examined during Stage 2 are excluded.

\begin{figure}[t]
  \centering
  \includegraphics[width=\columnwidth]{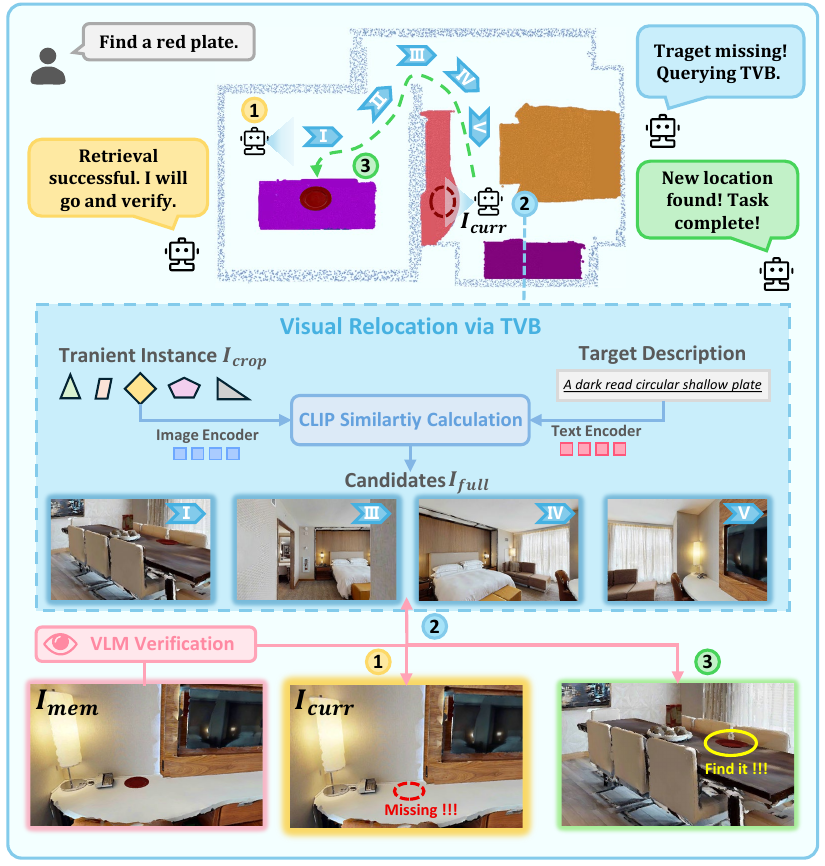}
  \caption{\textbf{Visual relocalization via Transient Visual Buffer (TVB).} If the retrieved target is absent from the current view, OASIS uses CLIP to match transient crops to the query and a VLM to verify candidate frames and confirm the target's new location.}
  \label{fig:nav}
\end{figure}

\subsection{Navigation and Relocation}
\label{sec:nav_reloc}

To close the perception-action loop, our system employs a view-driven navigation strategy integrated with a two-tier relocation mechanism, ensuring robustness against dynamic environmental changes.

\textbf{View-Driven Navigation \& Verification.} 
Upon retrieving the target instance $u_{target}$ in Sec.~\ref{sec:retrieval}, the system determines the optimal observation pose.
For targets identified via VLM (Stage 2), the frame explicitly marked as the \textit{best view} is selected; for those retrieved via CLIP fallback (Stage 3), the frame with the highest similarity score among all stored observations is chosen.
The robot then navigates to the recorded 6-DoF pose $T_{wc}$ of the selected frame, aligning its camera to replicate the original viewpoint.
Upon arrival, the robot captures a current observation $I_{curr}$. The VLM performs a visual comparison between the retrieved memory frame $I_{mem}$ and $I_{curr}$ to confirm the target's presence.

\textbf{Dynamic Object Relocation.} 
Should the verification fail (\ie, the object has moved), the system triggers the relocation procedure illustrated in Figure~\ref{fig:nav}. Since the robot does not perform dense mapping during navigation, we introduce a lightweight \textbf{Transient Visual Buffer} to capture potential sightings of the moved object.

\textit{1) Transient Instance Capture:} 
During navigation, we reuse the same object detection pipeline adopted in the mapping stage to process the video stream in real time. The matching step is performed against objects already stored in the semantic map; detections that cannot be matched to any mapped instance are flagged as ``Transient Instances''. For each such instance, we cache a tuple $\langle I_{crop}, I_{full}, T_{pose} \rangle$, storing the cropped object image, the full contextual frame, and the camera pose. To conserve memory, only the highest-confidence frame is retained for each transient track.

\textit{2) Visual Relocation:} 
When the target is reported missing, the system first queries this buffer. 
We compute CLIP similarities between the target's visual description (from Sec.~\ref{sec:retrieval}) and the cached $I_{crop}$ images, selecting the Top-5 candidates.
These 5 full frames, along with the original target image $I_{target}$, are fed into the VLM. The VLM is prompted to determine if the missing target appears in any of the new locations. 
Upon a positive match, the robot updates the target's position and navigates to the corresponding $T_{pose}$ to perform the final verification as described above.

\textit{3) Semantic Relocation:}
In scenarios where the visual search yields no matches (implying the object was not seen along the path), the system reverts to semantic reasoning. The LLM analyzes the OASIS graph again, excluding the current location, to predict the next most probable anchor (\eg, ``\textit{If the mug is not on the desk, check the coffee table nearby}''). The robot then navigates to the pose of the representative frame with the highest $S(\cdot)$ score for this predicted anchor and repeats the verification process.

\begin{table}[t]
    \centering
    \small
    \setlength{\tabcolsep}{8pt}
    \caption{\textbf{Fine-grained object retrieval.} Comparison on the OpenLex3D (Replica) benchmark.}
    \label{tab:replica_results}
    \begin{tabular}{lccc}
        \toprule
        Method & mAP $\uparrow$ & AP$_{50}$ $\uparrow$ & AP$_{25}$ $\uparrow$ \\
        \midrule
        ConceptGraphs~\cite{dense-1} & 2.74 & 6.18 & 13.82 \\
        HOV-SG~\cite{dense-2}        & 3.72 & 7.59 & 14.94 \\
        Kassab2024~\cite{kassab2024} & 0.66 & 2.15 & 5.34  \\
        OpenMask3D~\cite{CLIP-2} & 2.10 & 4.93 & 12.63 \\
        DualMap~\cite{light-2}       & 6.19 & 11.72 & 20.06 \\
        \textbf{Ours}                & \textbf{6.58} & \textbf{12.52} & \textbf{20.63} \\
        \bottomrule
    \end{tabular}
\end{table}
\section{Experiments}
\label{sec:experiments}

\subsection{Overview}
In our implementation, we use GPT-5.2 for LLM planning and Gemini 3 Pro for VLM verification. Since our main contribution is to improve semantic understanding and dynamic relocation capabilities, our evaluation across diverse retrieval and navigation tasks focuses on answering three key questions:
(1) \textbf{Complex Retrieval Capability:} Can OASIS accurately retrieve objects based on complex language queries?
(2) \textbf{Dynamic Adaptability:} Can the system effectively relocate objects that have moved after the initial mapping phase?
(3) \textbf{Component Efficacy:} What is the contribution of each module to the overall performance? Crucially, OASIS preserves the real-time efficiency required for online operation; detailed system profiling (\eg, mapping speed and inference latency) is provided in Appendices~\ref{supp:tpf} and~\ref{supp:full_dialogue}.

\subsection{Object Retrieval}
\label{subsec:exp_retrieval}

\subsubsection{Experimental Setup}
To comprehensively evaluate OASIS, we conduct experiments on three benchmarks spanning Replica~\cite{replica}, HM3D~\cite{hm3d}, and ScanNet~\cite{scannet} scenes, each targeting a different retrieval capability.

\noindent\textbf{1. Fine-grained Object Retrieval.}
We utilize the OpenLex3D benchmark~\cite{openlex} (Replica subset) to assess fine-grained object-level retrieval performance. Its human-annotated queries cover all scene objects and combine pure-synonym and synonym+descriptor forms, including fine-grained pattern, cultural, and brand cues (\eg, ``floral cup'', ``ironman portrait'', ``nike sneaker''), making the retrieval task more realistic and challenging.
\\
\textit{Metric:} Following OpenLex3D, we report mean Average Precision (mAP), as well as AP at IoU thresholds of 0.25 (AP$_{25}$) and 0.50 (AP$_{50}$).

\noindent\textbf{2. Functional Region Retrieval.}
To evaluate macro-level scene understanding, we follow the HOV-SG room retrieval protocol~\cite{dense-2} and its annotated room--object ground truth on eight HM3D scenes (15.6 regions/scene on average). This setup tests whether the model can accurately localize broad functional areas (\eg, ``kitchen'', ``office'') rather than isolated objects, under consistent task definitions and supervision.
\\
\textit{Metric:} Following HOV-SG, we report Top-5 accuracy over 20 frequent object categories; a query is successful if the predicted object overlaps the ground truth with IoU $>$ 0.1.

\begin{table}[t]
    \centering
    \small
    \setlength{\tabcolsep}{8pt}
    \caption{\textbf{Functional region retrieval.} Comparison on the HOV-SG (HM3D) benchmark.}
    \label{tab:room_retrieval}
    \begin{tabular}{lccc}
        \toprule
        Method & Regions & Trials & SR (\%) $\uparrow$ \\
        \midrule
        ConceptGraphs~\cite{dense-1} & \multirow{4}{*}{15.6} & \multirow{4}{*}{34.9} & 29.3 \\
        HOV-SG~\cite{dense-2}        &                        &                        & 31.5 \\
        DualMap~\cite{light-2}       &                        &                        & 53.8 \\
        \textbf{Ours}                &                        &                        & \textbf{57.7} \\
        \bottomrule
    \end{tabular}
\end{table}


\begin{table*}[t]
    \centering
    \small
    \setlength{\tabcolsep}{8pt}
    \caption{\textbf{Spatial relation retrieval.} Grounding accuracy on the Sr3D+ (ScanNet) benchmark across different difficulty splits.}
    \label{tab:sr3d_results}
    \resizebox{\textwidth}{!}{%
    \begin{tabular}{lcccccccccc}
        \toprule
        \multirow{2}{*}{Method} & \multicolumn{2}{c}{Overall (\%)} & \multicolumn{2}{c}{Easy (\%)} & \multicolumn{2}{c}{Hard (\%)} & \multicolumn{2}{c}{View-Dependent (\%)} & \multicolumn{2}{c}{View-Independent (\%)} \\
        \cmidrule(lr){2-3} \cmidrule(lr){4-5} \cmidrule(lr){6-7} \cmidrule(lr){8-9} \cmidrule(lr){10-11}
         & A@0.1 $\uparrow$ & A@0.25 $\uparrow$ & A@0.1 $\uparrow$ & A@0.25 $\uparrow$ & A@0.1 $\uparrow$ & A@0.25 $\uparrow$ & A@0.1 $\uparrow$ & A@0.25 $\uparrow$ & A@0.1 $\uparrow$ & A@0.25 $\uparrow$ \\
        \midrule
        OpenFusion~\cite{openfusion}       & 12.6 & 2.4 & 14.0 & 2.4 & 1.3 & 1.1 & 3.8 & 2.5 & 13.7 & 2.4 \\
        BBQ-CLIP~\cite{LLM-3}              & 14.4 & 8.8 & 15.4 & 9.0 & 6.7 & 6.7 & 11.4 & 5.1 & 14.4 & 8.8 \\
        ConceptGraphs~\cite{dense-1}       & 13.3 & 6.2 & 13.0 & 6.8 & 16.0 & 1.3 & 15.2 & 5.1 & 13.1 & 6.4 \\
        DualMap~\cite{light-2}             & 28.4 & 21.6 & 29.9 & 22.4 & 17.3 & 16.0 & 16.5 & 11.4 & 30.1 & 23.0 \\
        BBQ~\cite{LLM-3}                   & 34.2 & 22.7 & 34.3 & 22.7 & 33.3 & 22.7 & \textbf{32.9} & 20.3 & 34.4 & 23.0 \\
        CORE-3D~\cite{dense-3}             & 41.8 & 35.6 & 41.8 & 35.7 & \textbf{41.3} & \textbf{34.7} & \textbf{32.9} & \textbf{30.4} & 43.0 & 36.3 \\
        \textbf{Ours}                      & \textbf{44.0} & \textbf{37.4} & \textbf{44.7} & \textbf{37.9} & 38.7 & 33.3 & \textbf{32.9} & 29.1 & \textbf{45.5} & \textbf{38.5} \\
        \bottomrule
    \end{tabular}%
    }
\end{table*}

\begin{table}[t]
    \centering
    \caption{\textbf{Object Navigation Success Rate (SR) on HM3D.}}
    \label{tab:hm3d_nav_comparison}
    \small
    \setlength{\tabcolsep}{3.5pt}
    \resizebox{\columnwidth}{!}{%
    \begin{tabular}{llccccc}
        \toprule
        Scene Type & Method & 00829 & 00848 & 00880 & Trials & Avg. SR $\uparrow$ \\
        \midrule
        \multirow{4}{*}{Static} 
        & ConceptGraphs~\cite{dense-1} & 69.2 & 53.8 & 61.5 & \multirow{4}{*}{78} & 61.5 \\
        & HOV-SG~\cite{dense-2}        & 53.8 & 46.2 & 57.7 & & 52.6 \\
        & DualMap~\cite{light-2}       & 73.1 & 69.2 & 69.2 & & 70.5 \\
        & \textbf{Ours} & \textbf{76.9} & \textbf{92.6} & \textbf{84.6} & & \textbf{85.9} \\
        \midrule
        \multirow{2}{*}{In-Anchor} 
        & DualMap~\cite{light-2}       & 66.7 & 66.7 & 61.1 & \multirow{2}{*}{54} & 64.8 \\
        & \textbf{Ours} & \textbf{83.3} & \textbf{83.3} & \textbf{77.8} & & \textbf{81.5} \\
        \midrule
        \multirow{2}{*}{Cross-Anchor} 
        & DualMap~\cite{light-2}       & 55.6 & 61.1 & 64.7 & \multirow{2}{*}{53} & 60.3 \\
        & \textbf{Ours} & \textbf{77.8} & \textbf{83.3} & \textbf{70.6} & & \textbf{77.4} \\
        \bottomrule
    \end{tabular}
    }
\end{table}

\noindent\textbf{3. Spatial Relation Retrieval.}
The goal is to evaluate robust spatial relation retrieval under varying contextual complexities and viewpoint dependencies. We use the Sr3D+ benchmark~\cite{sr3d}, built on ScanNet scenes, and report four standard splits: \textit{Easy} ($\leq$2 distractors; \eg, ``The lamp closer to the white armchair.''), \textit{Hard} ($>$2 distractors; \eg, ``The lamp on the right in between the beds.''), \textit{View-Dependent} (needs a specific view; \eg, ``Looking at bed, the one on the left.''), and \textit{View-Independent} (viewpoint-free; \eg, ``The large cabinets directly above the fridge and oven.'').
\\
\textit{Metric:} Following CORE-3D~\cite{dense-3}, we report grounding accuracy at IoU thresholds of 0.1 and 0.25 (A@0.1, A@0.25).

\subsubsection{Results and Analysis}
As presented in Table~\ref{tab:replica_results}, our method achieves the best performance in \textbf{Fine-grained Object Retrieval} (6.58 mAP). The margin over DualMap is modest because both methods share the same instance-level scene construction pipeline. Our gain mainly comes from finer-grained attribute recognition, which yields higher feature similarity for the true target and pushes it to earlier ranks; this improves per-query AP and, consequently, leads to a slightly higher overall mAP. 
In \textbf{Functional Region Retrieval} (Table~\ref{tab:room_retrieval}), our method effectively handles this task. Although our scene graph has no explicit room-level hierarchy, our system determines functional areas by semantic reasoning over objects around each anchor together with visual verification of background layout. Table~\ref{tab:sr3d_results} details the \textbf{Spatial Relation Retrieval} results. Our method excels in the ``Easy'' and ``View-Independent'' splits, outperforming the specialized baseline CORE-3D. This confirms that our LLM planner effectively resolves local spatial constraints (\eg, ``next to''). In the ``Hard'' split, CORE-3D remains stronger, which suggests a natural boundary of our current design. These hard queries often involve long-range relationships (\eg, ``farthest from'') that span multiple anchors. Because each anchor representative frame usually captures objects near that anchor, our anchor-centric verification can be less effective when the required spatial relation is between far-apart objects.


\subsection{Navigation and Relocation}
\label{subsec:exp_navigation}

\subsubsection{Experimental Setup}
We evaluate closed-loop navigation performance on the HM3D dataset~\cite{hm3d}, adhering to the protocol established in DualMap~\cite{light-2}. The evaluation encompasses three scenarios: \textbf{Static}, where the environment remains unchanged; \textbf{In-Anchor Dynamic}, where target objects are moved within the same anchor region (\eg, shifted on a table); and \textbf{Cross-Anchor Dynamic}, where objects are relocated to entirely different rooms or distant anchors. We report Success Rate (SR), defined as the percentage of queries in which the agent stops within 1 meter of the queried object. In dynamic scenes, a query is counted as successful only if the target is found within three attempts.

\subsubsection{Results and Analysis}
Table~\ref{tab:hm3d_nav_comparison} summarizes the navigation results, where our method consistently outperforms all baselines. In the \textbf{Static} setting, our method achieves an impressive average Success Rate (SR) of 85.9\%, significantly surpassing DualMap (70.5\%) and ConceptGraphs (61.5\%). This substantial performance gain stems primarily from the superior retrieval accuracy of OASIS.
Our approach also shows clear advantages in dynamic scenarios. For \textbf{In-Anchor} moves, our method maintains a high SR of 81.5\%, demonstrating robustness to local object displacements. The most meaningful performance gain, however, is observed in the challenging \textbf{Cross-Anchor} setting, where our method achieves 77.4\% SR compared to DualMap's 60.3\%. DualMap adopts a strategy of continuous scene graph updating during navigation, which generally requires intentional scanning to capture small objects along the trajectory. By contrast, our Transient Visual Buffer (Sec. 3.5) passively captures transient object observations during robot transit, enabling efficient visual relocalization of displaced targets. Even if no new target location is detected along the navigation path, our semantic relocalization mechanism can further infer potential positions to support subsequent verification.

\begin{table}[t]
    \centering
    \caption{\textbf{Memory curation ablation.} Impact of different clustering and frame selection strategies on scene instance coverage.}
    \label{tab:ablation_mapping}
    \small
    \setlength{\tabcolsep}{5pt}
    \resizebox{\columnwidth}{!}{%
    \begin{tabular}{lcccc}
        \toprule
        Clustering & Global  & Local  & Adaptive & Avg. Clusters $\downarrow$ \\
        \midrule
        Spatial-Only     & 87.18 & 83.08 & 83.99 & 5.3 \\
        Visual-Only      & 84.79 & 72.92 & 86.64 & \textbf{3.8} \\
        \textbf{Hybrid (Ours)} & \textbf{90.82} & \textbf{89.30} & \textbf{91.13} & 6.5 \\
        \bottomrule
    \end{tabular}
    }
\end{table}

\subsection{Ablation Studies}
\label{sec:ablation}

\subsubsection{Memory Curation Strategy}
\label{sub:ablation_mapping}

We analyze the interplay between \textit{Clustering} and \textit{Frame Selection} using a full factorial design that combines three clustering methods (Spatial-Only, Visual-Only, Hybrid) with three selection scores (Global, Local, Adaptive Fusion). We report Instance Coverage, defined as the fraction of ground-truth Replica instances visible in at least one representative frame, and the average cluster count; all results are averaged over eight scenes.

As shown in Table~\ref{tab:ablation_mapping}, \textit{Visual-Only} clustering yields the fewest clusters but lower coverage because global CLIP features neglect spatial structure, causing confusion between distinct yet visually similar frames and different viewpoints. \textit{Spatial-Only} provides moderate coverage but cannot distinguish semantically different frames in close proximity. In contrast, \textit{Hybrid} achieves the highest coverage ($>$89\%) across all selection methods with a moderate cluster count. For frame selection, \textit{Local} underperforms and \textit{Global} remains stable, whereas \textit{Adaptive} reaches 91.13\% coverage, confirming the benefit of combining global context with local quality.

\subsubsection{Hierarchical Retrieval Pipeline}
\label{sub:ablation_retrieval}


We evaluate the contribution of each coarse-to-fine stage on Sr3D+ using three variants: (1) \textit{Full ablation} removes hierarchical reasoning and selects the highest-CLIP instance across the scene; (2) \textit{w/o LLM Proposal} obtains top-ranked instances through global CLIP matching, maps them to three unique anchors while skipping duplicates, and verifies their representative frames with the VLM; and (3) \textit{w/o VLM Verification} retains LLM-based anchor proposals but uses CLIP for final matching.

Table~\ref{tab:ablation_retrieval} confirms the necessity of both stages. Full ablation performs worst without spatial reasoning or fine-grained validation. The \textit{w/o LLM Proposal} variant retains reasonable accuracy but incurs high latency from global matching and unpruned anchor verification. The \textit{w/o VLM Verification} variant loses precision because CLIP cannot resolve fine-grained visual ambiguity or exact spatial relations. The full pipeline achieves the best accuracy--efficiency trade-off.


\begin{table}[t]
    \centering
    \caption{\textbf{Hierarchical retrieval ablation.} Impact of LLM proposal and VLM verification stages on the Sr3D+ benchmark.}
    \label{tab:ablation_retrieval}
    \small
    \setlength{\tabcolsep}{5pt}
    \begin{tabular}{cccc}
        \toprule
        LLM reasoning & VLM verification & A@0.1 (\%) $\uparrow$ & A@0.25 (\%) $\uparrow$ \\
        \midrule
        \ding{55} & \ding{55} & 28.4 & 21.6 \\ 
        \ding{55} & \ding{51} & 36.5 & 30.2 \\ 
        \ding{51} & \ding{55} & 32.8 & 26.9 \\ 
        \ding{51} & \ding{51} & \textbf{44.0} & \textbf{37.4} \\ 
        \bottomrule
    \end{tabular}
\end{table}
\begin{table}[t]
    \centering
    \caption{\textbf{Dynamic navigation ablation.} Impact of system components on relocalization performance in the Cross-Anchor setting.}
    \label{tab:ablation_navigation}
    \small
    \setlength{\tabcolsep}{4pt}
    \renewcommand{\arraystretch}{1.1}
    \resizebox{\columnwidth}{!}{%
    \begin{tabular}{lccc}
        \toprule
        Method & Attempts $\downarrow$ & SR (\%) $\uparrow$ & SPL (\%) $\uparrow$ \\
        \midrule
        Random Anchor & 2.9 & 17.0 & 3.9 \\
        Ours w/o Transient Visual Buffer & 2.5 & 56.6 & 17.3 \\
        Ours w/o Semantic Reasoning & 1.8 & 73.6 & 24.1 \\
        \textbf{Ours (Full)} & \textbf{1.6} & \textbf{77.4} & \textbf{25.8} \\
        \bottomrule
    \end{tabular}
    }
\end{table}

\subsubsection{Dynamic Navigation and Relocation}
\label{sub:ablation_navigation}

Finally, we assess navigation robustness in the \textit{Cross-Anchor} setting where objects are relocated to distant regions. We report Attempts (the number of relocation attempts per query) and SPL (success weighted by path length). We compare our full framework against a \textit{Random Anchor} baseline and two ablated variants: the baseline randomly navigates to the pose of the best representative frame of a random anchor, while the variant without Transient Visual Buffer performs semantic relocation at every step, and the one without Semantic Reasoning reverts to random anchor navigation if no new target location is detected in the buffer.

Quantitative results in Table~\ref{tab:ablation_navigation} demonstrate the superiority of our complementary design. The \textit{Random Anchor} baseline performs poorly with 17.0\% SR and 2.9 average attempts due to blind random search in the large scene space. Removing the visual buffer leads to limited performance (56.6\% SR) as pure semantic inference lacks reliable visual cues. Ablating semantic reasoning achieves moderate results (73.6\% SR) but suffers from inefficient random fallback. In contrast, our full framework integrates both components and achieves the best performance of 77.4\% SR with only 1.6 average attempts, validating the effectiveness of our dynamic relocalization mechanism.

\subsection{Real-World Experiments}
\label{subsec:real_world}
\begin{figure}[t]
    \centering
    \includegraphics[width=\columnwidth]{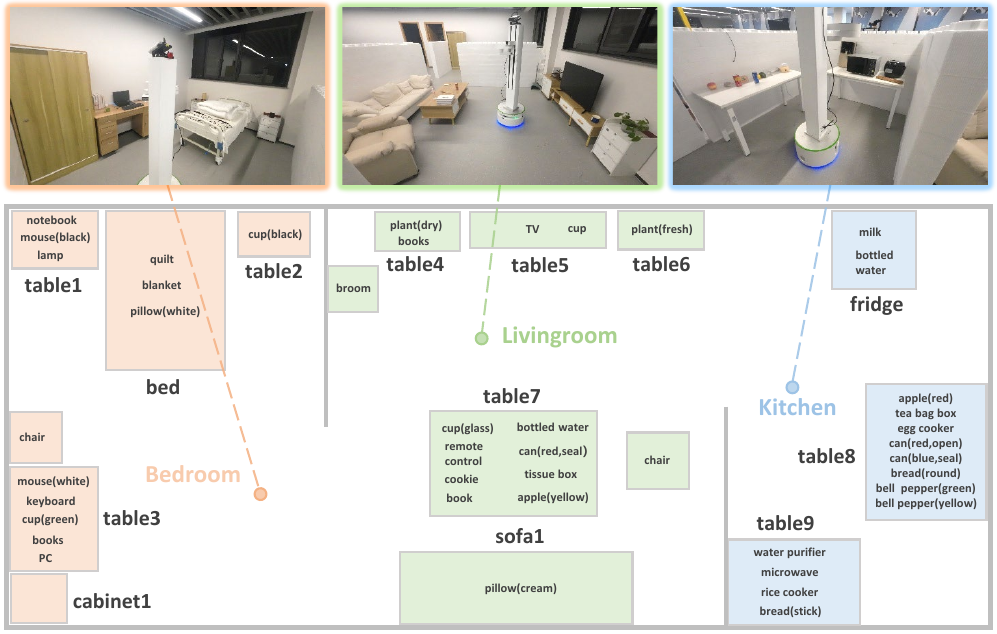}
    \caption{\textbf{Real-world household environment.} The three-room scene contains 13 semantic anchors and visually similar objects with diverse colors, shapes, and states.}
    \label{fig:realworld}
\end{figure}

As shown in Figure~\ref{fig:realworld}, we construct a real-world household environment comprising a bedroom, kitchen, and living room, with 13 semantic anchors. The scene includes visually similar objects with diverse categories, colors, shapes, and states, providing realistic distractors for open-vocabulary retrieval. We then conduct 30 retrieval-and-navigation trials in each of the static and dynamic settings. Static trials use fine-grained and relation-sensitive queries (\eg, \textit{Find a cabinet next to the TV with a dried potted plant on top}''). For dynamic trials, an LLM is given the scene layout and selects diverse targets and plausible relocation positions following common household habits.

As shown in Table~\ref{tab:realworld_static_dynamic}, OASIS outperforms DualMap in success rate and SPL while requiring fewer relocation attempts. Across 60 trials, we observe four static and seven dynamic failures. Static failures include two VLM attribute confusions (\eg, beige versus white pillows) and two mapping misses of small items (\eg, a remote control). Of the dynamic failures, four involve similar visual or mapping errors, two arise when distant-room searches are restricted to local anchors, and one results from a missed transient detection. Overall, perception and long-range reasoning remain the primary failure modes.

\begin{table}[t]
    \centering
    \caption{\textbf{Real-world navigation.} System performance under static and dynamic household settings.}
    \label{tab:realworld_static_dynamic}
    \small
    \setlength{\tabcolsep}{5pt}
    \resizebox{\columnwidth}{!}{%
    \begin{tabular}{lccccc}
        \toprule
        \multirow{2}{*}{Method} & \multicolumn{2}{c}{Static} & \multicolumn{3}{c}{Dynamic} \\
        \cmidrule(lr){2-3} \cmidrule(lr){4-6}
         & SR (\%) $\uparrow$ & SPL (\%) $\uparrow$ & SR (\%) $\uparrow$ & SPL (\%) $\uparrow$ & Attempts $\downarrow$ \\ 
        \midrule
        DualMap~\cite{light-2} & 63.3 & 58.2 & 56.7 & 18.1  & 2.1 \\
        Ours    & \textbf{86.7} & \textbf{76.3} & \textbf{76.7} & \textbf{27.6}  & \textbf{1.4} \\
        \bottomrule
    \end{tabular}
    }
\end{table}

\section{Conclusion}
We introduced OASIS, an online anchor-centric system for open-vocabulary retrieval and navigation in dynamic environments. Its multimodal memory and hierarchical LLM--VLM pipeline balance semantic expressiveness with online efficiency, while transient navigation observations support moved-object relocalization. Simulation and real-world experiments demonstrate improvements in retrieval accuracy, navigation success, and relocalization robustness over baselines, establishing anchor-centric memory as a practical foundation for long-horizon embodied intelligence.

\begin{acks}
This work was supported by the Shenzhen Science and Technology Program
[Grant No.~CJGJZD20240729141702003]. The authors would also like to thank
Shenzhen Pudu Technology Co., Ltd. for the support and resources provided,
which were essential to the successful completion of this research.
\end{acks}

\bibliographystyle{ACM-Reference-Format}
\bibliography{\OASISBibFile}

\clearpage
\appendix
\emergencystretch=1em
\setcounter{figure}{0}
\setcounter{table}{0}
\setcounter{equation}{0}
\renewcommand{\thefigure}{S\arabic{figure}}
\renewcommand{\thetable}{S\arabic{table}}
\renewcommand{\theequation}{S\arabic{equation}}
\renewcommand{\theHfigure}{supp.\arabic{figure}}
\renewcommand{\theHtable}{supp.\arabic{table}}
\renewcommand{\theHequation}{supp.\arabic{equation}}

\section{Supplementary Overview}
\label{supp:overview}

This supplementary material provides additional details that complement the main paper. Specifically, we include: (1) the prompt used for anchor-category definition; (2) the detailed formulation and hyperparameter settings of the scale-aware frame scoring module; (3) mapping memory/runtime overhead analysis; (4) the default Top-$K$ settings used in retrieval and relocalization; (5) an overview of the retrieval and relocalization pipeline together with stage-wise latency analysis; (6) prompt examples for the full retrieval--relocalization workflow.

\section{Anchor Definition Prompt}
\label{supp:prompt}

To determine the Anchor Object categories, we query the LLM with the predefined common indoor scene categories from DualMap~\cite{light-2} and ask it to select only the categories suitable as anchors.

\begin{quote}
\small\raggedright
\textbf{System Instruction:}
You are an expert in Spatial AI and Indoor Navigation. Your task is to identify which categories can serve as ``Anchor Objects'' from a given category list.

\textbf{Input:}
A Python list of category names from DualMap's predefined common indoor scene categories.

\textbf{Selection Criteria:}
\begin{enumerate}
    \item \textbf{Large Volume:} The object must have a significant physical footprint.
    \item \textbf{Immovable:} The object is heavy or structurally fixed (e.g., furniture that is rarely moved).
    \item \textbf{Cognitive Landmark:} Humans naturally use this object to locate other smaller items (e.g., ``The keys are on the \textit{table}'').
\end{enumerate}

\textbf{Negative Constraints (Do NOT include):}
\begin{itemize}
    \item Pure background surfaces such as wall, floor, and ceiling.
    \item Small movable furniture (e.g., dining chair, stool).
    \item Any category not present in the provided category list.
\end{itemize}

\textbf{Output Format:}
Return strictly a Python list of strings containing only selected anchor categories from the input list, in lowercase.
Do not add new categories that are not in the input.
Example output format: \texttt{['bed', 'sofa',} \texttt{'wardrobe', 'refrigerator']}
\end{quote}

Using this prompt, our anchor-category output includes: sofa, table, bed, desk, cabinet, shelf, television stand, refrigerator, oven, sink, washing machine, dryer, stove, dishwasher, toilet, shower, bathtub, door, window, air conditioner, heater, dresser, nightstand, wardrobe, closet, coat rack, staircase, and drawer.

\section{Detailed Formulation of Frame Scoring}
\label{supp:scoring}

In this section, we provide the detailed mathematical formulation and hyperparameter settings for the scale-aware frame scoring module described in the main paper.

\subsection{Scale-Aware Adaptive Weights}
Given $M$ anchor objects in a frame, let $\text{area}_i$ denote the pixel area of the $i$-th anchor. We first define the normalized anchor-area distribution as
\begin{equation}
    p_i = \frac{\text{area}_i}{\sum_{j=1}^{M} \text{area}_j}.
\end{equation}
The anchor entropy is then computed as
\begin{equation}
    H_{\text{anchor}} = -\sum_{i=1}^{M} p_i \log(p_i).
\end{equation}
This term measures the balance of anchor areas: higher entropy indicates that multiple anchors contribute more evenly.

To further capture the absolute visual footprint of anchors, we define the anchor coverage ratio
\begin{equation}
    r = \frac{\sum_{i=1}^{M} \text{area}_i}{A_{\text{img}}},
\end{equation}
where $A_{\text{img}}$ is the image area.

Based on these two cues, the adaptive weights are defined as
\begin{equation}
    w_{\text{global}} =
    \begin{cases}
    1-r, & M \le 1,\\
    (1-\lambda)(1-r) + \lambda \dfrac{H_{\text{anchor}}}{\log M}, & M > 1,
    \end{cases}
\end{equation}
\begin{equation}
    w_{\text{local}} = 1 - w_{\text{global}},
\end{equation}
where $\lambda \in [0,1]$ balances coverage and entropy, and is empirically set to 0.7. For $M \le 1$, anchor balance is not meaningful, so the weighting reduces to the coverage term alone.

\subsection{Global and Local Terms}
Both global and local terms are normalized to $[0,1]$.

\textbf{Global Term.}
The raw global score is a linear combination of anchor count and object count, clipped by upper bounds $A_{\max}$ and $O_{\max}$:
\begin{equation}
    S_{\text{raw}}^{\text{global}} = \alpha \cdot \min(M, A_{\max}) + (1-\alpha) \cdot \min(N_{\text{obj}}, O_{\max}).
\end{equation}
The normalized global score is
\begin{equation}
    S_{\text{global}} = \frac{S_{\text{raw}}^{\text{global}}}{\alpha A_{\max} + (1-\alpha) O_{\max}}.
\end{equation}

\textbf{Local Term.}
For each anchor $i$, we compute a quality value from edge completeness $E_i \in [0,1]$ and the clipped interaction count
\begin{equation}
    V_i = \beta \cdot E_i + (1-\beta) \cdot \min(C_i, C_{\max}).
\end{equation}
We then aggregate these values using both the maximum and mean across anchors:
\begin{equation}
    S_{\text{raw}}^{\text{local}} =
    \begin{cases}
        \gamma \cdot \max_i(V_i) + (1-\gamma) \cdot \dfrac{1}{M}\sum_i V_i, & M > 0,\\
        0, & M = 0.
    \end{cases}
\end{equation}
Finally, we normalize the local term by its maximum possible value:
\begin{equation}
    S_{\text{local}} = \frac{S_{\text{raw}}^{\text{local}}}{\beta + (1-\beta) C_{\max}}.
\end{equation}
Since $E_i \in [0,1]$ and $\min(C_i, C_{\max}) \in [0, C_{\max}]$, the maximum possible value of $V_i$ is $\beta + (1-\beta) C_{\max}$, so this normalization ensures $S_{\text{local}} \in [0,1]$.

\subsection{Hyperparameters and Sensitivity Analysis}
The default hyperparameters used in our experiments are listed in Table~\ref{tab:params}.

\begin{table}[h]
    \centering
    \caption{Hyperparameters for frame scoring.}
    \label{tab:params}
    \begin{tabular*}{\columnwidth}{@{\extracolsep{\fill}} l c p{0.56\columnwidth} @{} }
    \toprule
    Parameter & Value & Description \\
    \midrule
    $\lambda$ & 0.7 & {\footnotesize Balance weight between coverage and entropy} \\
    $\alpha$ & 0.7 & {\footnotesize Weight for anchor count in the global score} \\
    $A_{\max}, O_{\max}$ & 5.0, 10.0 & {\footnotesize Saturation thresholds for counts} \\
    $\beta$ & 0.5 & {\footnotesize Weight for completeness in the local score} \\
    $C_{\max}$ & 5.0 & {\footnotesize Saturation threshold for interactions} \\
    $\gamma$ & 0.7 & {\footnotesize Weight for max aggregation in local score fusion} \\
    \bottomrule
    \end{tabular*}
\end{table}

To study the sensitivity of the proposed frame scoring scheme, we perform a one-factor-at-a-time analysis by varying one hyperparameter while fixing all others to their default values in Table~\ref{tab:params}. We evaluate $\lambda$, $\alpha$, $\beta$, and $\gamma$ over the same range of values, and report scene instance coverage as the evaluation metric. For the complementary parameter pairs, we use $1-\alpha$, $1-\beta$, and $1-\gamma$ as the corresponding weights for object count, interaction richness, and mean aggregation, respectively.

\begin{figure}[t]
    \centering
    \includegraphics[width=\columnwidth]{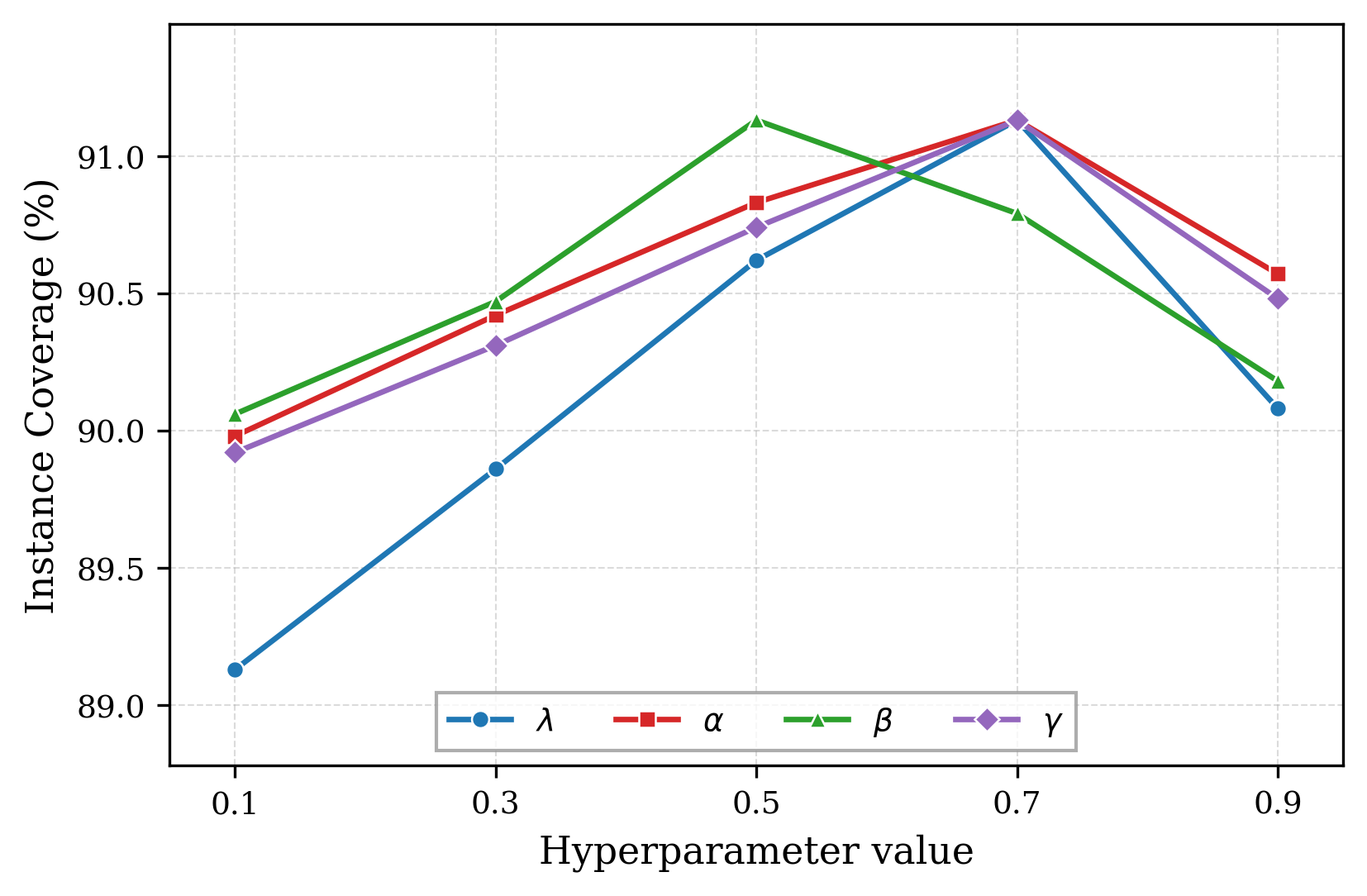}
    \caption{Sensitivity analysis of frame-scoring hyperparameters.}
    \label{fig:hyper_sensitivity}
\end{figure}

Fig.~\ref{fig:hyper_sensitivity} shows that the proposed method remains stable across a broad range of hyperparameter settings. In particular, the coverage curves vary only slightly when changing $\lambda$, $\alpha$, $\beta$, or $\gamma$, indicating that the scoring function is reasonably robust around the default configuration. Although the absolute best value may vary slightly, the adopted defaults consistently fall within strong-performing regions. We therefore use $\lambda=0.7$, $\alpha=0.7$, $\beta=0.5$, and $\gamma=0.7$ in all experiments.

\section{Mapping Overhead Analysis}
\label{supp:tpf}

We report memory and runtime overhead for ConceptGraphs~\cite{dense-1}, HOV-SG~\cite{dense-2}, DualMap~\cite{light-2}, and our method. These measurements are collected during the online mapping stage and exclude LLM/VLM calls used only at retrieval time. All measurements are conducted on a single NVIDIA RTX 4090 GPU. TPF denotes the average processing time per RGB-D frame. Specifically, we compare average memory (Avg.\ Mem), peak memory (Peak Mem), and time per frame (TPF). Our method is built on top of DualMap and adds a parallel episodic memory curation module, resulting in only a marginal increase in memory and time overhead. As shown in Table~\ref{tab:tpf_extracted}, this added cost is minimal while retaining practical real-time performance and remaining far more efficient than dense methods.

\begin{table}[t]
    \centering
    \caption{Memory and runtime overhead comparison.}
    \label{tab:tpf_extracted}
    \small
    \setlength{\tabcolsep}{3pt}
    \resizebox{\columnwidth}{!}{%
    \begin{tabular}{l l c c c}
        \toprule
        Dataset & Method & Avg. Mem (MB) $\downarrow$ & Peak Mem (MB) $\downarrow$ & TPF (s) $\downarrow$ \\
        \midrule
        \multirow{4}{*}{Replica} & ConceptGraphs & 7148.9 & 23551.9 & 4.188 \\
         & HOV-SG & 73368.0 & 158126.6 & 42.005 \\
         & DualMap & \textbf{3095.2} & \textbf{4564.0} & \textbf{0.276} \\
         & Ours & 3231.2 & 4971.1 & 0.312 \\
        \midrule
        \multirow{4}{*}{ScanNet} & ConceptGraphs & 9780.3 & 26155.2 & 6.301 \\
         & HOV-SG & 9223.0 & 25735.0 & 8.039 \\
         & DualMap & \textbf{2120.9} & \textbf{2820.2} & \textbf{0.163} \\
         & Ours & 2237.6 & 3164.3 & 0.198 \\
        \bottomrule
    \end{tabular}
    }
\end{table}

\section{Default Top-\texorpdfstring{$K$}{K} Settings}
\label{supp:topk}

Our system uses three default Top-$K$ settings during retrieval and relocalization.

\textbf{Top-$3$ anchor candidates.}
In the LLM-based anchor proposal stage, we keep the top-$3$ candidate anchors for subsequent verification. This choice provides a practical balance between search-space reduction and retrieval robustness: using too few anchors may prematurely discard plausible target regions, while using too many anchors increases VLM calls and overall latency.

\textbf{Top-$5$ representative frames per anchor.}
For each anchor, we retain at most $5$ representative frames ranked by frame score. This setting is intended to preserve diverse yet high-quality observations of the same anchor without introducing excessive redundancy. In practice, a small set of representative views is sufficient to cover common viewpoint variations around stable indoor anchors.

\textbf{Top-$5$ transient visual buffer (TVB) relocation candidates.}
During dynamic relocalization, we first rank TVB entries by CLIP similarity and keep the top-$5$ candidate full frames for VLM verification. This design limits the verification cost while preserving enough diversity to recover relocated objects from transient observations along the navigation path.

Unless otherwise specified, these Top-$K$ values are fixed throughout all experiments. They were selected based on preliminary development experience to provide a reasonable trade-off between robustness, memory usage, and inference latency.

\section{Retrieval and Relocalization Example and Latency Analysis}
\label{supp:full_dialogue}

To improve reproducibility, we provide an overview of one full example session from retrieval to dynamic relocalization. The detailed prompt example is provided in Sec.~\ref{supp:prompt_screenshots}.

\subsection{Prompt Example Overview}

\noindent \textbf{Stage 1: LLM-based anchor proposal prompt.}

In Stage 1, the LLM takes the user query and structured anchor metadata as input, then proposes and ranks candidate anchors by relevance to the target object. The prompt explicitly constrains the output format and anchor-ID validity, enabling deterministic downstream parsing.

\noindent \textbf{Stage 2: anchor-based VLM verification prompt.}

In Stage 2, the VLM inspects representative frames from the selected anchors and performs fine-grained verification to determine whether the target can be visually confirmed. The output includes completion status, best frame, and target attributes (anchor ID, relation, class, and compact visual description).

\noindent \textbf{Stage 3: post-navigation target existence verification prompt.}

After navigation and camera rotation toward the previously confirmed target location, Stage 3 verifies whether the target still exists in the current view. If not found, the model reports disappearance and provides a comparison-based explanation between reference and current observations.

\noindent \textbf{Stage 4: TVB-based candidate re-identification prompt.}

When the object is missing at the original location, Stage 4 queries the TVB and re-verifies candidate views for relocalization. Specifically, the original images of the candidate instances are compared with the target reference image to select the best-matching relocated candidate.

\noindent \textbf{Stage 5: same-instance relocalization verification prompt.}

In Stage 5, the agent navigates to the newly retrieved location and performs a second verification pass to confirm successful dynamic object relocalization. This prompt explicitly checks whether the current view matches the previously retrieved candidate and whether the target can be confirmed as the same instance at the relocated position.

This example illustrates the complete loop: anchor proposal, fine-grained target verification, disappearance detection, TVB query, and final dynamic object relocalization verification.

\subsection{Latency Analysis}

\begin{table}[htbp]
    \centering
    \caption{Latency of LLM/VLM interactions across stages.}
    \label{tab:stage_latency}
    \begin{tabular*}{\columnwidth}{@{\extracolsep{\fill}} l l c c @{} }
        \toprule
        Stage & API & Avg. Image Count & Mean Latency (s) \\
        \midrule
        Stage 1 & GPT-5.2 & -- & 4.2 \\
        Stage 2 & Gemini-3-Pro & 6.4 & 14.9 \\
        Stage 3 & Gemini-3-Pro & 2.0 & 8.1 \\
        Stage 4 & Gemini-3-Pro & 6.0 & 12.7 \\
        Stage 5 & Gemini-3-Pro & 2.0 & 7.6 \\
        \bottomrule
    \end{tabular*}
\end{table}

To provide a comprehensive understanding of system performance during deployment, we report a stage-wise latency breakdown of the hierarchical retrieval pipeline. Table~\ref{tab:stage_latency} shows the average time spent on cloud-based LLM/VLM interactions across the five stages of a typical open-vocabulary retrieval and relocalization session. As shown in Table~\ref{tab:stage_latency}, the end-to-end latency is currently dominated by cloud API inference and network I/O (using GPT-5.2 and Gemini-3-Pro), rather than by the local graph operations themselves. In our implementation, local operations such as scene-graph traversal, node filtering, and TVB updates execute in milliseconds. Thus, the current bottleneck mainly reflects the use of offboard multimodal reasoning models. 

Importantly, this latency does not interfere with low-level robot control. Our architecture decouples high-level semantic reasoning from low-level motion execution: the API-induced delay occurs during the reasoning and verification phase, while the underlying navigation stack runs locally on the compact scene graph. Moreover, passive observations for the TVB are processed locally during movement without requiring heavy VLM calls, ensuring uninterrupted navigation. Therefore, the reported latency should be understood primarily as a deployment artifact of cloud-based multimodal models rather than a structural limitation of the OASIS framework. As edge-side multimodal models continue to improve, the same pipeline can be migrated to onboard deployment with substantially lower reasoning latency.

\clearpage
\onecolumn
\captionsetup{hypcap=false}
\section{Detailed Prompt Example}
\label{supp:prompt_screenshots}

This section provides the detailed prompt example for the user query \textit{``Find a red plate.''} corresponding to the five-stage end-to-end retrieval-and-relocalization workflow summarized in Sec.~\ref{supp:full_dialogue}, where Stage~1 corresponds to Fig.~\ref{fig:prompt1}, Stage~2 to Fig.~\ref{fig:prompt2}, Stage~3 to Fig.~\ref{fig:prompt3}, Stage~4 to Fig.~\ref{fig:prompt4}, and Stage~5 to Fig.~\ref{fig:prompt5}. Note that these five stages describe the full deployment workflow and therefore extend the three retrieval stages described in the main paper.

{\centering
\includegraphics[width=\textwidth,height=0.79\textheight,keepaspectratio]{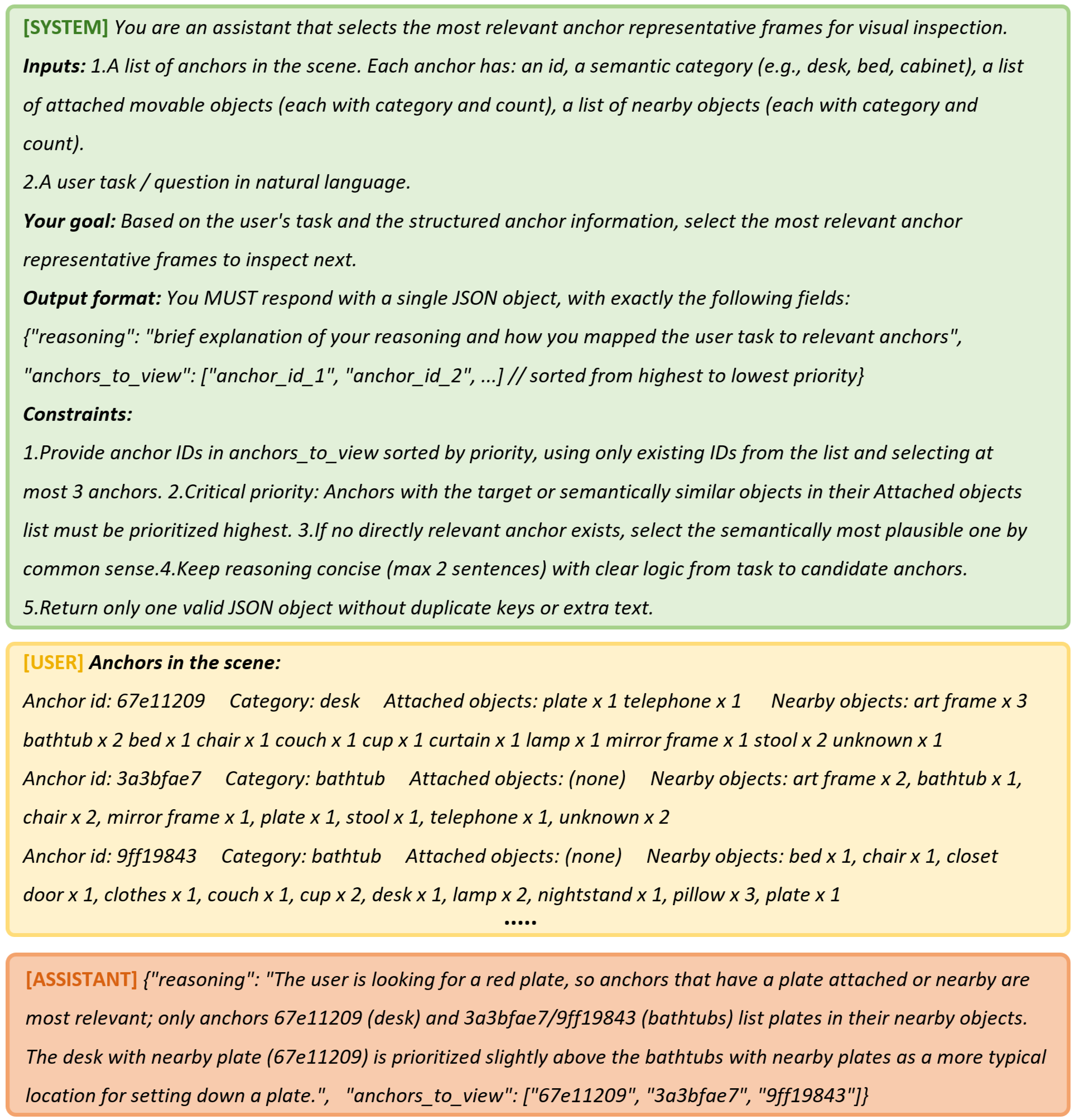}
\captionof{figure}{Stage 1: LLM-based anchor proposal prompt.}
\label{fig:prompt1}\par}

\clearpage
{\centering
\includegraphics[width=\textwidth,height=0.79\textheight,keepaspectratio]{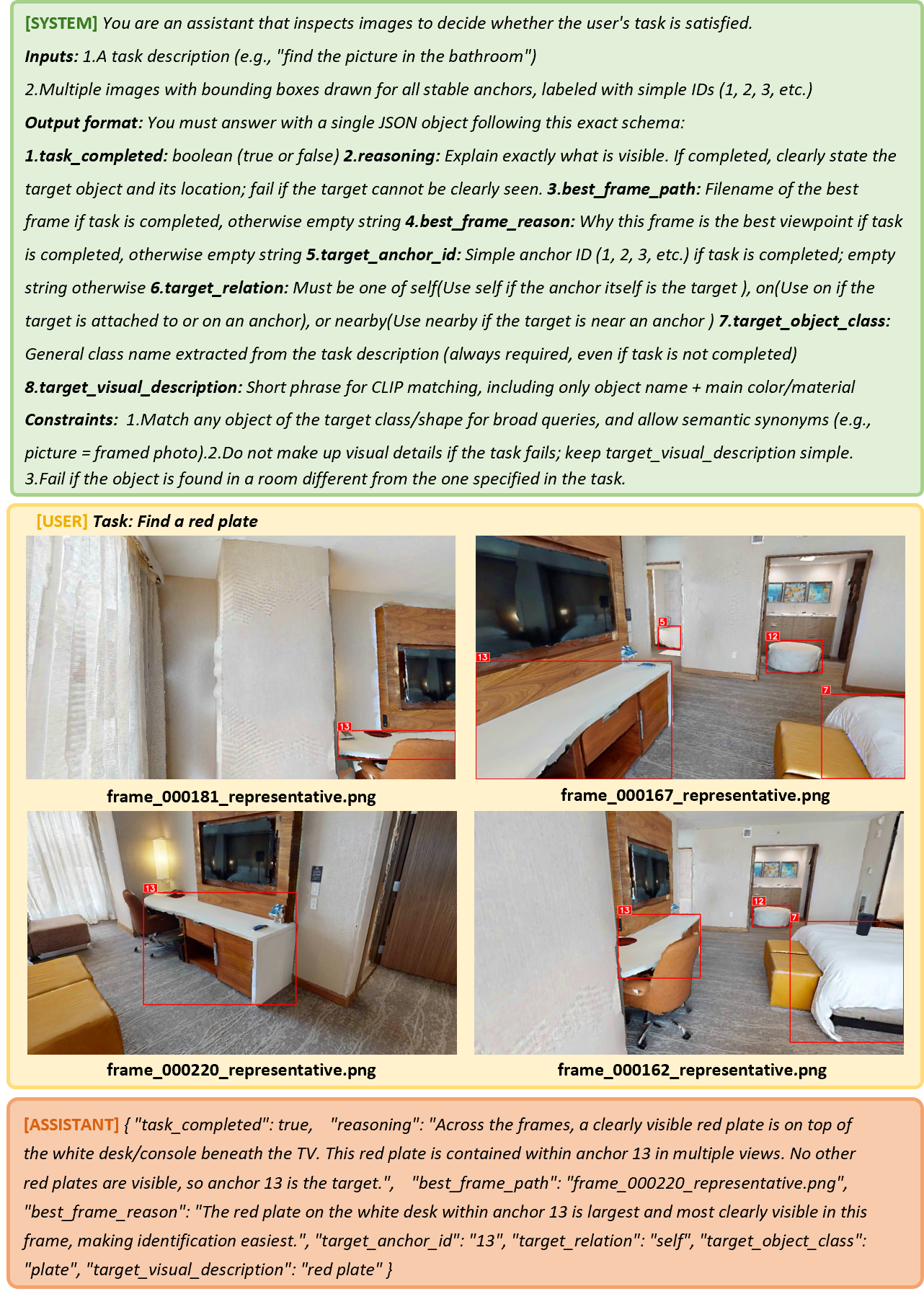}
\captionof{figure}{Stage 2: anchor-based VLM verification prompt.}
\label{fig:prompt2}\par}

\clearpage
{\centering
\includegraphics[width=\textwidth,height=0.79\textheight,keepaspectratio]{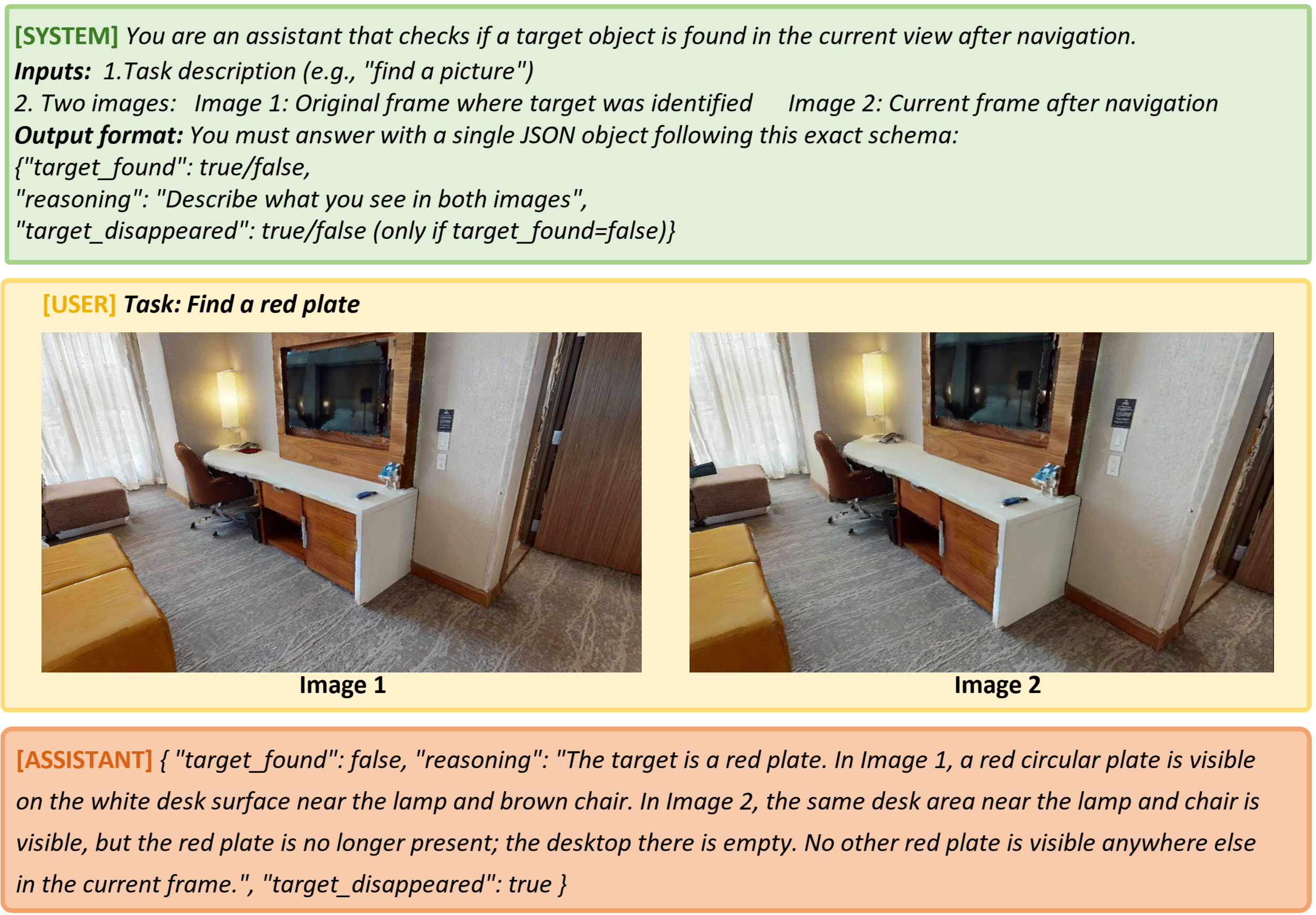}
\captionof{figure}{Stage 3: post-navigation target existence verification prompt.}
\label{fig:prompt3}\par}

\clearpage
{\centering
\includegraphics[width=\textwidth,height=0.79\textheight,keepaspectratio]{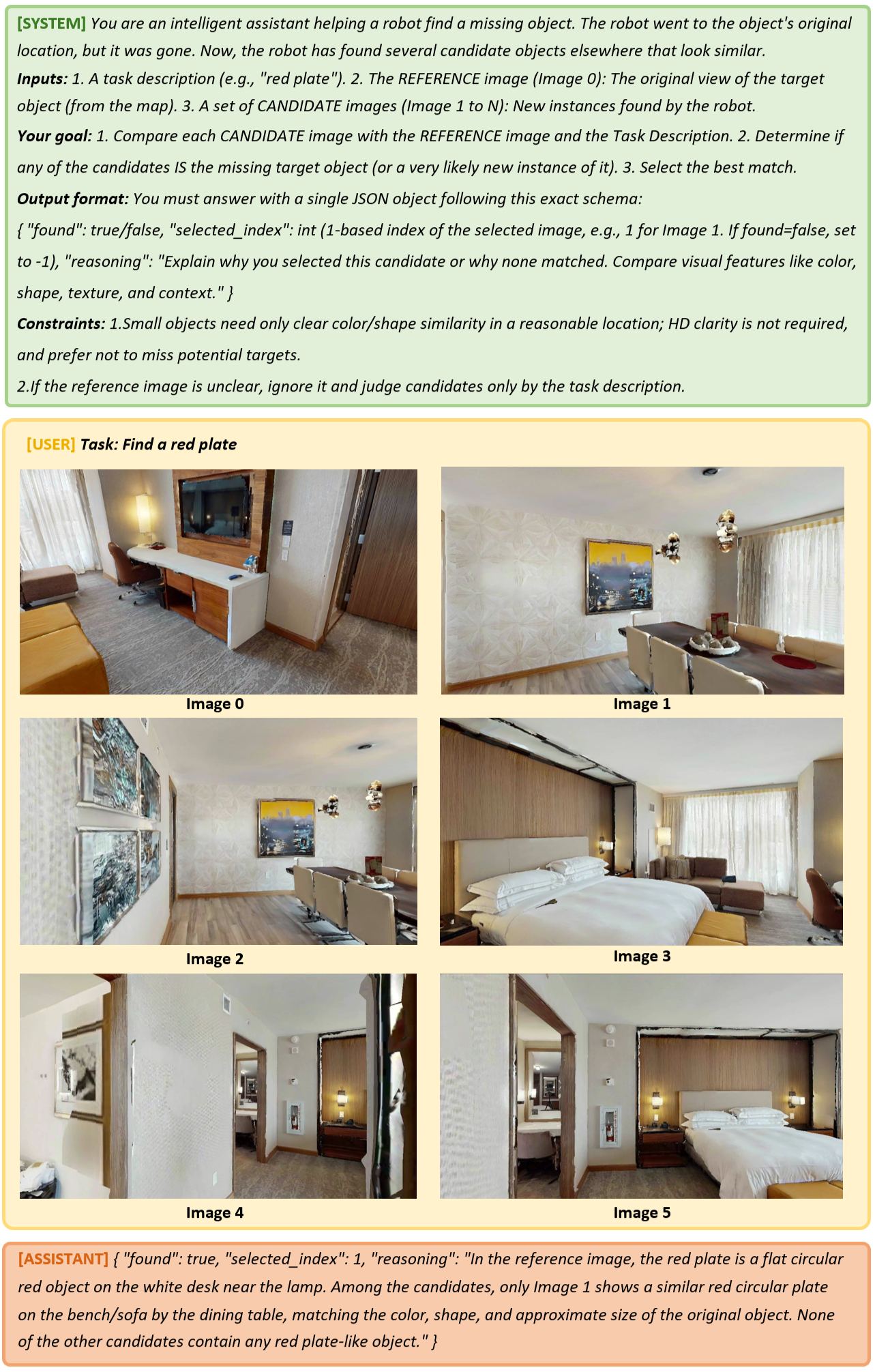}
\captionof{figure}{Stage 4: TVB-based candidate re-identification prompt.}
\label{fig:prompt4}\par}

\clearpage
{\centering
\includegraphics[width=\textwidth,height=0.79\textheight,keepaspectratio]{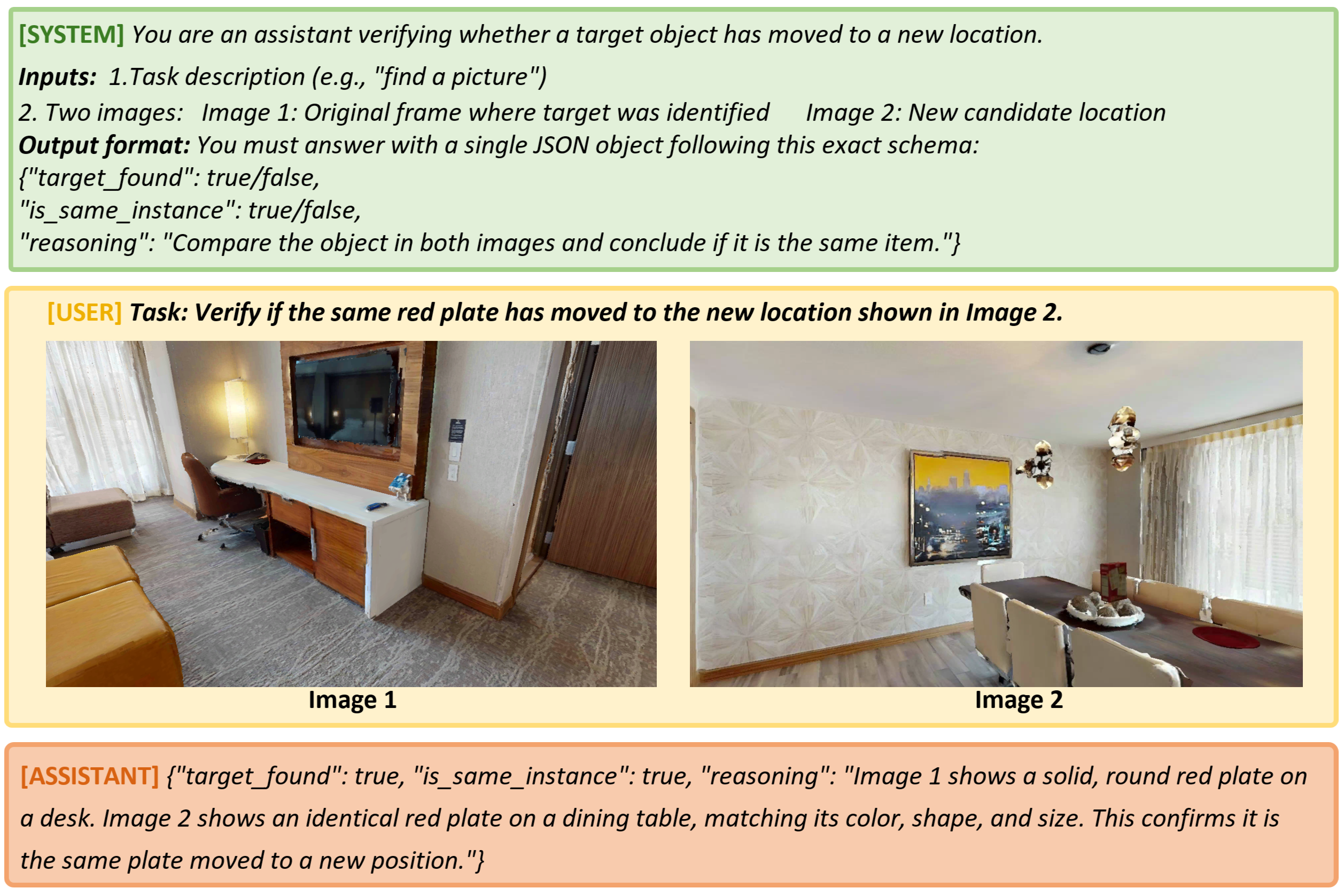}
\captionof{figure}{Stage 5: same-instance relocalization verification prompt.}
\label{fig:prompt5}\par}

\end{document}